\documentclass[sn-apa,iicol]{sn-jnl}

\usepackage{mathrsfs}%

\usepackage[title]{appendix}%
\usepackage{textcomp}%
\usepackage{manyfoot}%
\usepackage{listings}%
\usepackage[caption=false]{subfig}
\usepackage{multicol}
\usepackage{graphicx}

\makeatletter
\@ifpackageloaded{xcolor}{}{%
\usepackage[table,x11names,dvipsnames,svgnames]{xcolor}%
}
\makeatother

\usepackage{colortbl}

\usepackage{graphicx}
\usepackage{wrapfig}

\definecolorset{RGB}{lyft}{}{Red,194,39,36;Sunset,202,53,33;Orange,205,68,20;Amber,200,117,42;Yellow,242,169,52;Citron,186,188,44;Lime,112,159,33;Green,56,139,31;Mint,45,118,56;Teal,52,133,135;Cyan,60,132,202;Blue,55,94,248;Indigo,64,13,247;Purple,115,42,248;Pink,176,25,145;Rose,176,32,75}

\usepackage{suffix}

\usepackage{environ}

\makeatletter
\newsavebox{\boxifnotempty}
\newcommand{\displayifnotempty}[3]{\sbox\boxifnotempty{#2}\setbox0=\hbox{\usebox{\boxifnotempty}\unskip}%
\ifdim\wd0=0pt
\else
 #1\usebox{\boxifnotempty}#3%
\fi%
}
\newcommand{\ifempty}[2]{\setbox0=\hbox{#1\unskip}%
\ifdim\wd0=0pt%
 #2%
\fi%
}
\newcommand{\ifnotempty}[2]{\setbox0=\hbox{#1\unskip}%
\ifdim\wd0>0pt%
 #2%
\fi%
}
\makeatother

\usepackage{algpseudocode}
\usepackage{algorithm}

\usepackage{scrlfile}

\makeatletter
\newcommand*\newstoreddef[1]{
  \BeforeClosingMainAux{%
    \immediate\write\@auxout{%
      \string\restoredef{#1}{\csname #1\endcsname}%
    }%
  }%
}
\newcommand*{\restoredef}[2]{
  \expandafter\gdef\csname stored@#1\endcsname{#2}%
}
\newcommand*{\storeddef}[1]{
  \@ifundefined{stored@#1}{0}{\csname stored@#1\endcsname}%
}
\makeatother

\usepackage{pageslts}
\NewEnviron{tee}{\BODY\typeout{Marker Tee [start] ^^J \BODY ^^JMaker Tee [end]}}

\input{preamble/math}
\newcommand{\real}[1]{\mathbb{R}^{#1}{}}

\newcommand{\bmat}[1]{\begin{bmatrix}#1\end{bmatrix}}

\newcommand{\smallbmat}[1]{\left[\begin{smallmatrix}#1\end{smallmatrix}\right]}

\DeclareMathOperator*{\argmin}{\arg\!\min}

\providecommand{\cH}{\mathcal{H}}

\usepackage{units}

\usepackage{microtype}

\usepackage{array}
\usepackage{multirow}
\usepackage{booktabs}
\usepackage{makecell} 

\ifcsname labelindent\endcsname

\fi
\usepackage[inline]{enumitem}

\newenvironment{lenumerate}[2][]
{\begin{enumerate}[label=(#2\arabic*),leftmargin=0.2in,itemindent=0.15in,#1]}
{\end{enumerate}}

\setlist*[enumerate,1]{label={\itshape\arabic*)}}

\makeatletter
\newcommand{\paragraphswithstop}{%
\let\copyparagraph\paragraph%
\renewcommand\paragraph[1]{\copyparagraph{##1.}}%
}
\makeatother

\usepackage[framemethod=tikz]{mdframed}

\usepackage{tikz}
\usetikzlibrary{calc}
\usetikzlibrary{matrix}
\usetikzlibrary{chains}
\usetikzlibrary{shapes.geometric}
\usetikzlibrary{arrows.meta}
\usetikzlibrary{decorations.pathreplacing}
\usetikzlibrary{backgrounds}

\tikzset{
  dim above/.style={to path={\pgfextra{
        \pgfinterruptpath
        \draw[>=latex,|->|] let
        \p1=($(\tikztostart)!1.5em!90:(\tikztotarget)$),
        \p2=($(\tikztotarget)!1.5em!-90:(\tikztostart)$)
        in(\p1) -- (\p2) node[pos=.5,sloped,above]{#1};
        \endpgfinterruptpath
      }
    }
  },
  dim double above/.style={to path={\pgfextra{
        \pgfinterruptpath
        \draw[>=latex,|->|] let
        \p1=($(\tikztostart)!3em!90:(\tikztotarget)$),
        \p2=($(\tikztotarget)!3em!-90:(\tikztostart)$)
        in(\p1) -- (\p2) node[pos=.5,sloped,above]{#1};
        \endpgfinterruptpath
      }
    }
  },
  dim below/.style={to path={\pgfextra{
        \pgfinterruptpath
        \draw[>=latex,|->|] let 
        \p1=($(\tikztostart)!-1em!-90:(\tikztotarget)$),
        \p2=($(\tikztotarget)!-1em!90:(\tikztostart)$)
        in (\p1) -- (\p2) node[pos=.5,sloped,below]{#1};
        \endpgfinterruptpath
      }
    }
  },
}

\tikzset{
    right angle quadrant/.code={
        \pgfmathsetmacro\quadranta{{1,1,-1,-1}[#1-1]}     
        \pgfmathsetmacro\quadrantb{{1,-1,-1,1}[#1-1]}},
    right angle quadrant=1, 
    right angle length/.code={\def\rightanglelength{#1}},   
    right angle length=2ex, 
    right angle symbol/.style n args={3}{
        insert path={
            let \p0 = ($(#1)!(#3)!(#2)$) in     
                let \p1 = ($(\p0)!\quadranta*\rightanglelength!(#3)$), 
                \p2 = ($(\p0)!\quadrantb*\rightanglelength!(#2)$) in 
                let \p3 = ($(\p1)+(\p2)-(\p0)$) in  
            (\p1) -- (\p3) -- (\p2)
        }
    }
}

\newcommand{\pgfextractangle}[3]{%
    \pgfmathanglebetweenpoints{\pgfpointanchor{#2}{center}}
                              {\pgfpointanchor{#3}{center}}
    \global\let#1\pgfmathresult  
}

\usetikzlibrary{shapes.arrows}

\tikzset{ax/.style={-latex,line width=2pt}}

\tikzset{camera/.style={fill=Sienna1,fill opacity=0.5},%
image plane/.style={draw=RoyalBlue3,line width=2pt}}

\usepackage{cleveref}
\usepackage[final]{changes}
\setdeletedmarkup{\textcolor{red}{\sout{#1}}}
\setaddedmarkup{\textcolor{red}{\uline{#1}}}

\begin{document}

\title[Article Title]{Learning Personalized Safety Interventions for Haptic Human-Robot Shared Control}

\author*[1]{\fnm{Dawei} \sur{Zhang}}\email{dwzhang@bu.edu}
\author[1]{\fnm{Roberto} \sur{Tron}}\email{tron@bu.edu}

\affil[1]{\orgdiv{Division of Systems Engineering}, \orgname{Boston University}, \orgaddress{\street{110 Cummington Mall}, \city{Boston}, \postcode{02215}, \state{MA}, \country{USA}}}

\abstract{
Haptic feedback provides an implicit channel for communicating safety intentions during human-robot shared control.  Existing haptic guidance systems typically employ predefined intervention strategies that cannot accommodate the diverse safety preferences of individual users or application scenarios. To address this limitation, we propose a Learning from Haptics (LfH) framework that learns user-preferred safety interventions from sparse demonstrations, eliminating the need for manual trial-and-error design. Our framework is built on a differentiable Control Barrier Function (CBF)-based optimization layer that automatically adjusts the underlying safety parameters to match the demonstrated haptic responses. Instead of tuning controller parameters directly, users teach the system how they expect it to intervene during teleoperation. The resulting haptic guidance reflects the demonstrated intervention preferences while preserving the intuitive interaction of haptic shared control. Simulation and hardware experiments demonstrate that the proposed framework can learn personalized safety interventions from sparse user input and reduce the mismatch between the generated haptic feedback and the demonstrated preferences.}


\maketitle

\section{Introduction}\label{sec1}

Teleoperation enables human operators to perform tasks in hazardous or unstructured environments where fully autonomous operation remains challenging. In such scenarios, haptic shared control has emerged as an effective paradigm for improving safety while preserving the operator's decision-making authority by communicating safety-related information through force feedback~\citep{Lam2009,Brandt2010,Hou2013}. Recently, Control Barrier Functions (CBFs) have been adopted to generate haptic safety interventions for shared-control teleoperation~\citep{auro2024,cbf_vf2026}. Rather than directly modifying the operator's command, a virtual CBF-based safety controller computes a constraint-compliant reference, and the discrepancy between this reference and the operator's command is rendered as haptic force feedback. In this way, the haptic interface provides an implicit communication channel for the underlying safety intervention while allowing the human operator to retain full control authority over the robot.

Although CBFs provide a principled mechanism for generating safety interventions, the resulting haptic behavior is largely determined by the response parameters of the underlying CBF. These parameters influence both when the intervention begins and how strongly the corrective guidance is
communicated to the operator. More broadly, recent CBF studies have shown that selecting or adapting
class-$\mathcal{K}$ functions and their associated response parameters can substantially affect constraint feasibility, conservativeness, and closed-loop behavior
\citep{zeng2021optimaldecay,xiao2022adaptive, parwana2025ratetunable,chriat2023optimality}. Consequently, different parameter values can produce substantially different intervention behaviors even for the same teleoperation task. For example, an operator may prefer earlier and stronger
interventions when approaching pedestrians, while preferring weaker guidance near static obstacles. Such preferences depend not only on the application scenario but also on the individual operator.

Existing haptic shared control (HSC) systems typically rely on manually tuned response parameters. However, obtaining a desired intervention behavior through manual tuning is difficult because the mapping from the optimization parameters to the rendered haptic feedback is implicitly defined by the solution of the CBF-based quadratic program (CBF-QP). As a result, even small parameter adjustments can lead to unintuitive changes in the resulting haptic behavior, making manual tuning time-consuming and requiring considerable expertise.

To address these limitations, we propose a Learning from Haptics (LfH) framework that allows users to specify preferred haptic interventions through sparse demonstrations. Rather than learning an end-to-end control policy, the framework treats the CBF-QP as an implicit model of the intervention and adapts its response parameters through differentiable optimization. This preserves the original shared-control structure and the operator's control authority while enabling user-specific haptic safety guidance.

The main contributions of this work are summarized as follows:

\begin{itemize}

\item We propose an LfH framework that enables users to personalize optimization-based safety interventions for HSC through sparse demonstrations of their preferred force responses.

\item We formulate personalized safety-intervention learning as a differentiable optimization problem by embedding the CBF-QP into the learning pipeline. This allows the response parameters to be learned directly from user demonstrations while retaining the structured CBF-based intervention mechanism.

\item We validate the proposed framework through hardware-in-the-loop simulation and real-world experiments, demonstrating personalized haptic safety interventions across different safety representations using the same learning framework.

\end{itemize}

\subsection{Related Work}
\label{sec:related_work}
Haptic feedback has long been recognized as an effective means of communicating navigation and safety information during robot teleoperation. Early approaches generated guidance forces from explicitly designed safety metrics, including parametric risk fields (PRFs)~\citep{Lam2009}, time-to-impact (TTI)~\citep{Brandt2010}, and dynamic kinesthetic boundaries (DKBs)~\citep{Hou2013}. Similar ideas have been applied to aerial robots, mobile robots, and powered wheelchairs to improve operator situational awareness and collision avoidance \citep{Morere2015HapticAssistance,VanderPoorten2012PoweredFeedback,Coffey2023ReactiveGuidance}. More recently, Control Barrier Functions (CBFs) have provided a principled framework for generating haptic safety interventions by rendering the difference between the operator's command and a safety-compliant reference as force feedback~\citep{zhang2020haptic,zhang2021haptic}. These methods preserve human control authority while implicitly communicating safety information through haptic interaction. However, the intervention behavior is determined by manually selected controller parameters and therefore cannot adapt to different users or application scenarios.

Learning has also been introduced into human-robot interaction to improve teleoperation performance and personalize user interfaces. Existing studies have learned personalized mappings from human motion or body-machine interfaces to robot commands~\citep{macchini2020personalized}, while others have investigated haptic feedback as a mechanism for improving operator perception and motor learning during simulated drone teleoperation~\citep{rognon2019haptic}. These approaches demonstrate that learning can improve the usability and personalization of teleoperation. However, their learning objectives primarily concern the human-to-robot command mapping or operator skill acquisition. In contrast, our objective is to preserve the original shared-control paradigm and learn how an optimization-based safety controller should communicate its intervention through haptic feedback.

A growing body of work combines learning with CBFs for safety-critical control. Existing approaches have learned uncertain dynamics to reduce the conservativeness of safety filters~\citep{wang2018safe,taylor2020learning}, learned barrier functions or safe sets from demonstrations and neural representations~\citep{Robey2020,long2021learning,so2024train}, and integrated CBFs with reinforcement learning to ensure safe exploration or policy execution~\citep{cheng2019end,chen2019enhancing}. These methods use learning
to improve safety certificates, identify uncertain system components, or optimize autonomous control policies. In contrast, we assume that the safety controller is already available and instead learn the response parameters that
govern the haptic safety intervention perceived by the human operator.

Recent studies have also investigated how the choice of class-$\mathcal{K}$ functions affects the behavior and feasibility of CBF-based controllers. Optimal-decay and adaptive CBF formulations adjust decay rates or penalty functions to improve QP feasibility under input constraints and uncertainty \citep{zeng2021optimaldecay,xiao2022adaptive}. Rate-tunable CBFs
parameterize the class-$\mathcal{K}$ functions and adapt their parameters online to regulate the approach toward the safe-set boundary \citep{parwana2025ratetunable}, while learning-based approaches optimize
parameterized class-$\mathcal{K}$ functions together with autonomous control policies \citep{chriat2023optimality}. These studies show that the CBF response is not determined by the safe set alone, but can be shaped through the associated class-$\mathcal{K}$ functions. Our work builds on this observation by learning the response gains from haptic demonstrations, so that the resulting intervention reflects how the
operator prefers safety guidance to be communicated.

Recent advances in differentiable optimization have made it possible to integrate optimization-based controllers directly into learning architectures. OptNet introduced quadratic programs as differentiable neural network layers~\citep{amos2017optnet}, while differentiable convex optimization layers generalized this idea to parameterized convex optimization problems~\citep{agrawal2019differentiable}. BarrierNet further incorporated differentiable CBF-QPs into learning architectures to jointly optimize control policies and safety-related parameters~\citep{Xiao2023BarrierNet,liu2023barrier}. Inspired by these developments, we treat the CBF-QP as an implicit model of the haptic safety intervention and differentiate through its solution to learn the CBF response parameters directly from sparse user demonstrations. Unlike existing learning-enabled CBF approaches, our framework does not learn a barrier function, system dynamics, or an autonomous policy; instead, it learns personalized safety interventions while preserving the original HSC framework and human control authority.

\section{Preliminaries}\label{sec:pre}
This section reviews the key concepts used in the proposed framework, including Control Barrier Function (CBF)-based haptic shared control (HSC) and Gaussian Process (GP) regression.

\subsection{Control Barrier Functions}
\label{sec:review_cbf}

This subsection briefly reviews the CBF formulation used in this paper. Since
the proposed HSC framework considers acceleration-controlled robot dynamics,
we focus on second-order systems and the corresponding exponential CBF
formulation.

Consider the input-affine dynamical system
\begin{equation}
\label{state model}
    \dot{x}=f(x)+g(x)u,
\end{equation}
where $x\in\real{n}$ is the system state, $u\in\real{d}$ is the control input,
$f:\real{n}\rightarrow\real{n}$, and
$g:\real{n}\rightarrow\real{n\times d}$.
The vector fields $f$ and the columns of $g$ are assumed to be locally
Lipschitz~\citep{Ames2019}.

For the aerial teleoperation system considered in this work, the robot is
modeled as a double integrator, with acceleration as the control input. Let
$x=\smallbmat{x_p\\x_v}\in\real{2d}$, where
$x_p\in\real{d}$ denotes the position and
$x_v=\dot{x}_p\in\real{d}$ denotes the velocity. The dynamics are given by
\begin{equation}
\label{eq:double integrator}
    \bmat{\dot{x}_p\\\dot{x}_v}
    =
    \bmat{0&I\\0&0}
    \bmat{x_p\\x_v}
    +
    \bmat{0\\I}u,
\end{equation}
where $I$ is the identity matrix of appropriate dimension.
\subsubsection{Lie derivatives}

For a continuously differentiable function $h(x)$, its Lie derivative along
the vector field $f(x)$ is denoted by
\[
L_fh(x)=\dot{h}=\frac{\partial h(x)}{\partial x}f(x).
\]
Throughout this paper, we assume that $h(x)$ has relative degree two with
respect to the system dynamics, i.e., $L_gh=0$ and
$L_gL_fh\neq0$. Consequently, $\ddot h=L_f^2h+L_gL_fhu$.


\subsubsection{Safety Set}
We define a safe set $\cH$ (e.g., the area \emph{not} occupied by an obstacle) as the zero superlevel set of a continuously differentiable function $h(x)$:
\begin{equation*}\label{eqn:safety_set}
\cH: = \left\{ x \in\real{n} : h (x)\geq 0 \right\}.
\end{equation*}
The set $\cH$ is said to be \emph{forward invariant} for system \eqref{state model} if every solution starting from any $x(t_0)\in \cH$ satisfies $x(t)\in \cH$ for all $t \geq t_0$.

\subsubsection{CBFs for Second-Order Systems}

For the double-integrator dynamics in
\eqref{eq:double integrator}, we assume that the safety function $h(x)$ has
relative degree two with respect to the system dynamics. Safety is enforced
using an exponential CBF~\citep{Nguyen2016}, which imposes the constraint
\begin{equation}
\label{eq:HOCBF}
    \ddot{h}
    \geq
    -\alpha_1(h)
    -\alpha_2(\dot h),
\end{equation}
where $\alpha_1(\cdot)$ and $\alpha_2(\cdot)$ are class-$\mathcal{K}$
functions. Throughout this paper, we use linear class-$\mathcal{K}$ functions,
\[
\alpha_1(h)=k_1h,\qquad
\alpha_2(\dot h)=k_2\dot h,
\]
where $k_1,k_2>0$ are referred to as the CBF response gains.
Substituting the double-integrator dynamics into
\eqref{eq:HOCBF} yields the following linear constraint on the control input:
\begin{equation}
\label{eq:CBF_constraint}
\tag{CBF constraints}
x_v^\top \partial_{x_p}^2 h\,x_v
+
(\partial_{x_p}h)^\top u
+
k_2(\partial_{x_p}h)^\top x_v
+
k_1 h
\geq 0.
\end{equation}

The response gains $k_1$ and $k_2$ determine the dynamic response of the CBF.
Specifically, $k_1$ weights the clearance term $h$, while $k_2$ weights the
approach-rate term $\dot h$, jointly influencing when the safety intervention
begins and how aggressively it responds as the system approaches the boundary
of the safe set. 

The CBF constraint is enforced through the following quadratic program (QP),
which computes the control input that minimally deviates from a reference
control while satisfying the safety constraint:
\begin{equation}
\label{eq:CBF_QP}
\begin{aligned}
u_{\mathrm{cbf}}
=
\arg\min_{u\in\mathbb{R}^d}
\quad &
\frac{1}{2}\|u-u_{\mathrm{ref}}\|^2\\
\mathrm{s.t.}\quad
&
\eqref{eq:CBF_constraint}.
\end{aligned}
\end{equation}

This formulation naturally extends to environments containing multiple obstacles by introducing one CBF constraint for each safety set. If applied directly to the robot, the solution of
\eqref{eq:CBF_QP} corresponds to the standard CBF safety filter and provides the associated safety guarantees under the required feasibility assumptions. In the HSC paradigm, however, $u_{\text{cbf}}$ is used only as a virtual safety command for generating haptic force feedback, while the operator's command remains the actual control input applied to the robot.

\subsection{CBF-based Force Feedback}\label{sec:force feedback}
 We design the force feedback $F$ perceived by the user as the difference between a user-dependent reference control $u_{\text{ref}}$ and the safe control $u_{\text{cbf}}$ returned by the CBF-QP, i.e.,
\begin{equation}\label{eq:force}
    F_{\text{cbf}} = k_{\text{f}}(u_{\text{cbf}} - u_{\text{ref}}),
\end{equation}
where $k_{\text{f}}>0$ is a constant parameter to adjust the magnitude of the feedback. The optimal $u_{\mathrm{cbf}}$ is used only to generate $ F_{\text{cbf}}$, whereas the actual control applied to the
robot remains $u_{\mathrm{ref}}$. Thus, the operator retains direct control authority over the robot.
Following~\citep{zhang2020haptic}, the reference control $u_{\text{ref}}$ is defined as:
\begin{equation}\label{eq:uref}
    u_{\textrm{ref}} = \frac{1}{\Delta_t}(x_{vd}-x_v),
\end{equation}
where $x_{vd}$ is the desired velocity set by the user through the haptic input interface, $x_v$ is the current velocity of the robot,  and $\Delta_t$ is a time constant representing how long $u_{\textrm{ref}}$ will be applied to the robot; intuitively, this control aims to drive $x_v$ to $x_{vd}$ in a period of duration $\Delta_t$. 

\begin{remark}
In this scheme, the interaction between the user and the system through the input $x_{vd}$ and the output $F$ represents a cyberphysical feedback loop that might generate instabilities if the behavior of the user is too aggressive. 
To address this potential instability, the proposed framework is also applicable to formulations augmented with
$\mathcal{L}_2$-gain stability constraints
\citep{zhang2021stable,auro2024}. However, in the present paper, these terms are omitted to simplify
the presentation and focus on the learning framework.
\end{remark}

\subsection{Gaussian Process}
\label{sec:GP}

In the proposed framework, the user specifies preferred haptic feedback only at
a sparse set of time instants during replay. To recover a continuous desired
force trajectory for learning, we employ Gaussian Process (GP)
regression~\citep{rasmussen2006gaussian} to interpolate these sparse
annotations (as discussed in detail in \Cref{subsec:data_collection}). GP regression provides a principled probabilistic framework for
interpolating sparse observations while preserving smoothness of the resulting
trajectory.

A Gaussian Process defines a distribution over functions $f(x)$ such that the
function values evaluated at any finite set of input points follow a joint
Gaussian distribution with mean function $\mu(x)$ and covariance function
$k(x,x')$. Given observations
$y_1=f(X_1)$ at input locations $X_1$, GP regression predicts the posterior
distribution of
$y_2=f(X_2)$ at another set of input locations $X_2$:
\begin{align}
    p(y_2 \mid y_1, X_1, X_2)
    &= \mathcal{N}(\mu_{2|1}, \Sigma_{2|1}),\\
    \mu_{2|1}
    &= \mu_2
    + \Sigma_{21}\Sigma_{11}^{-1}(y_1-\mu_1),\\
    \Sigma_{2|1}
    &= \Sigma_{22}
    - \Sigma_{21}\Sigma_{11}^{-1}\Sigma_{12},
\end{align}
where $\mu_1 = \mu(X_1)$, $\mu_2 = \mu(X_2)$, $\Sigma_{11}= k(X_1,X_1)$, $\Sigma_{21}= k(X_2,X_1)$, and $\Sigma_{22}= k(X_2,X_2)$.

In this work, $X_1$ denotes the sparse time instants at which the user provides
preferred force annotations, while $X_2$ corresponds to the remaining samples
along the recorded teleoperation trajectory. The posterior mean
$\mu_{2|1}$ defines the interpolated desired haptic force trajectory used for
learning. The corresponding data collection and interpolation procedure is
described in Section~\ref{subsec:data_collection}.

\section{Problem Formulation}\label{sec:problem}
\begin{figure*}
    \centering
    \includegraphics[width=1\linewidth]{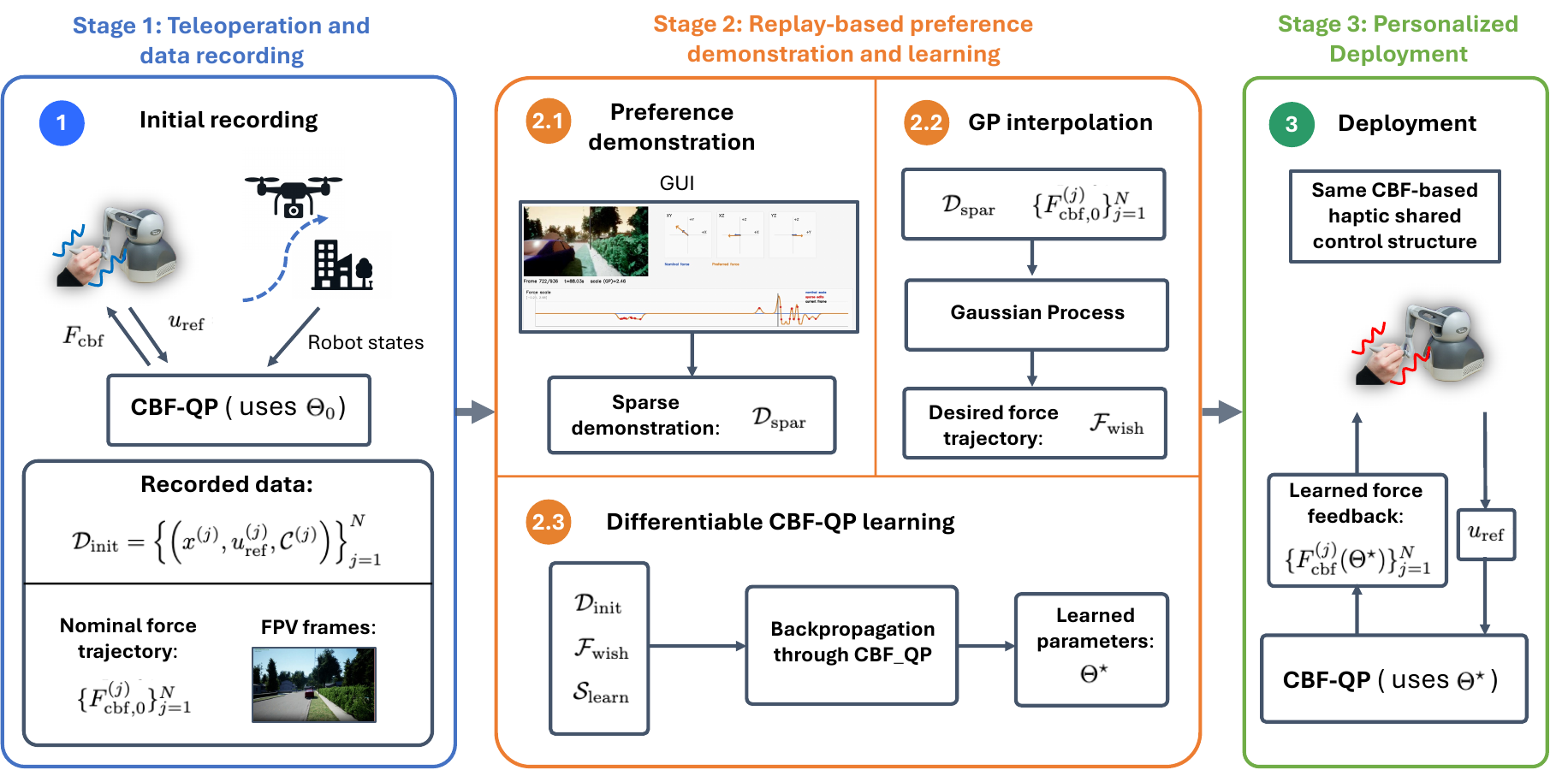}
    \caption{{\textbf{Overview of the proposed Learning from Haptics framework.} The pipeline proceeds from initial teleoperation and data recording to replay-based preference specification, differentiable CBF-QP learning, and deployment with personalized response gains.}}
    \label{fig:framework}
\end{figure*}

The proposed Learning from Haptics framework is summarized in
\Cref{fig:framework}. It consists of three stages: teleoperation and data
recording, replay-based preference specification and parameter learning,
and deployment with the learned CBF response gains. During the initial
teleoperation session, the operator specifies the desired velocity through
the haptic interface, from which the reference input $u_{\mathrm{ref}}$
is computed as described in \Cref{sec:force feedback}. Given the current
robot state and active safety constraints, the CBF-QP in
\eqref{eq:CBF_QP} computes $u_{\mathrm{cbf}}$, and the difference between
$u_{\mathrm{cbf}}$ and $u_{\mathrm{ref}}$ is rendered as the haptic force
$F_{\mathrm{cbf}}$. The robot remains under the operator's control, while
the CBF-QP output is used only for haptic feedback generation.

During this session, the robot state, reference input, active constraint
data, generated haptic force, and first-person-view (FPV) observations are
recorded. In the replay-based preference specification stage, the operator
reviews the recorded session and sparsely modifies the force feedback
according to the desired intervention behavior. These modifications are
interpolated to construct a desired force trajectory. The recorded data
and desired force trajectory are then used to learn the CBF response gains
through differentiable optimization. The learned gains are subsequently
deployed in the same CBF-based HSC framework to generate personalized
haptic interventions.

We consider a recorded teleoperation trajectory containing $N$ samples.
At sample $j$, the available quantities are the robot state $x^{(j)}$,
the reference input $u_{\mathrm{ref}}^{(j)}$, and the constraint data
$\mathcal{C}^{(j)}$. The recorded data used for learning are denoted by
\begin{equation}
\mathcal{D}_{\mathrm{init}}
=
\left\{
\left(
x^{(j)},
u_{\mathrm{ref}}^{(j)},
\mathcal{C}^{(j)}
\right)
\right\}_{j=1}^{N},
\label{eq:recorded_data}
\end{equation}
where $\mathcal{C}^{(j)}$ contains the geometric and group-related
quantities required to reconstruct the active CBF constraints at sample
$j$.

Let $\mathcal{S}$ denote the set of constraint groups, and let
$s_i^{(j)}\in\mathcal{S}$ denote the group assigned to the $i$-th active
constraint at sample $j$. Each group $s$ is associated with the CBF
response gains $\theta_s=[k_{1,s},\,k_{2,s}]^\top$. The learnable
parameter vector is
\begin{equation}
\Theta
=
\left\{
\theta_s
\right\}_{s\in\mathcal{S}_{\mathrm{learn}}},
\qquad
\mathcal{S}_{\mathrm{learn}}\subseteq\mathcal{S},
\label{eq:learnable_parameters}
\end{equation}
while the remaining groups retain their nominal gains.

For a given $\Theta$, let $u_{\mathrm{cbf}}^{(j)}(\Theta)$ denote the solution of \eqref{eq:CBF_QP} evaluated using $\mathcal{D}_{\mathrm{init}}$. The corresponding
generated haptic force is
\begin{equation}
F_{\mathrm{cbf}}^{(j)}(\Theta)
=
k_f
\left(
u_{\mathrm{cbf}}^{(j)}(\Theta)
-
u_{\mathrm{ref}}^{(j)}
\right).
\label{eq:parameterized_force}
\end{equation}
Let $\Theta_0$ denote the nominal CBF response gains used during the initial teleoperation session, and let
$F_{\mathrm{cbf},0}^{(j)}= F_{\mathrm{cbf}}^{(j)}(\Theta_0)$ denote the corresponding nominal haptic force at sample $j$. The nominal force trajectory is denoted by $\{ F_{\mathrm{cbf},0}^{(j)}\}_{j=1}^{N}$.

The replay-based preference specification procedure (described in
\Cref{subsec:data_collection}) constructs the desired force trajectory
$\mathcal{F}_{\mathrm{wish}}
=
\{F_{\mathrm{wish}}^{(j)}\}_{j=1}^{N}$, where
$F_{\mathrm{wish}}^{(j)}$ denotes the intervention preferred by the
operator for the recorded state, reference input, and active constraints
at sample $j$.

The personalized response gains $\Theta^\star$ are obtained by minimizing the discrepancy between the generated and desired force trajectories. Because the generated force is implicitly determined by the solution of the CBF-QP, the response gains are optimized by backpropagating through a differentiable CBF-QP layer. The learning objective and the corresponding differentiation procedure are presented in \Cref{subsubsec:learning_objective}
and \Cref{subsubsec:differentiable_optimization}, respectively.

\noindent
\textbf{Problem Statement.}
Given the recorded trajectory $\mathcal{D}_{\mathrm{init}}$, the desired
force trajectory $\mathcal{F}_{\mathrm{wish}}$, and the learnable
constraint groups $\mathcal{S}_{\mathrm{learn}}$, determine the response
gains $\Theta^\star$ such that the generated force trajectory
$\{F_{\mathrm{cbf}}^{(j)}(\Theta^\star)\}_{j=1}^{N}$ matches the
operator's demonstrated preference while preserving the CBF-based HSC
structure.

\section{Methodology}\label{sec:methods}
This section presents the methodology used to solve the personalized haptic intervention learning problem formulated in \Cref{sec:problem}. We first describe how an initial teleoperation session is recorded and how sparse user modifications are converted into the desired force trajectory $\mathcal{F}_{\mathrm{wish}}$. We then present the differentiable CBF-QP formulation used to optimize the response gains $\Theta$. Finally, we describe how the active CBF constraints are constructed from the environment representations used in the simulation and hardware experiments.

\subsection{Data Collection and Preparation}
\label{subsec:data_collection}

\begin{figure*}[!t]
    \centering
    \includegraphics[width=\linewidth]{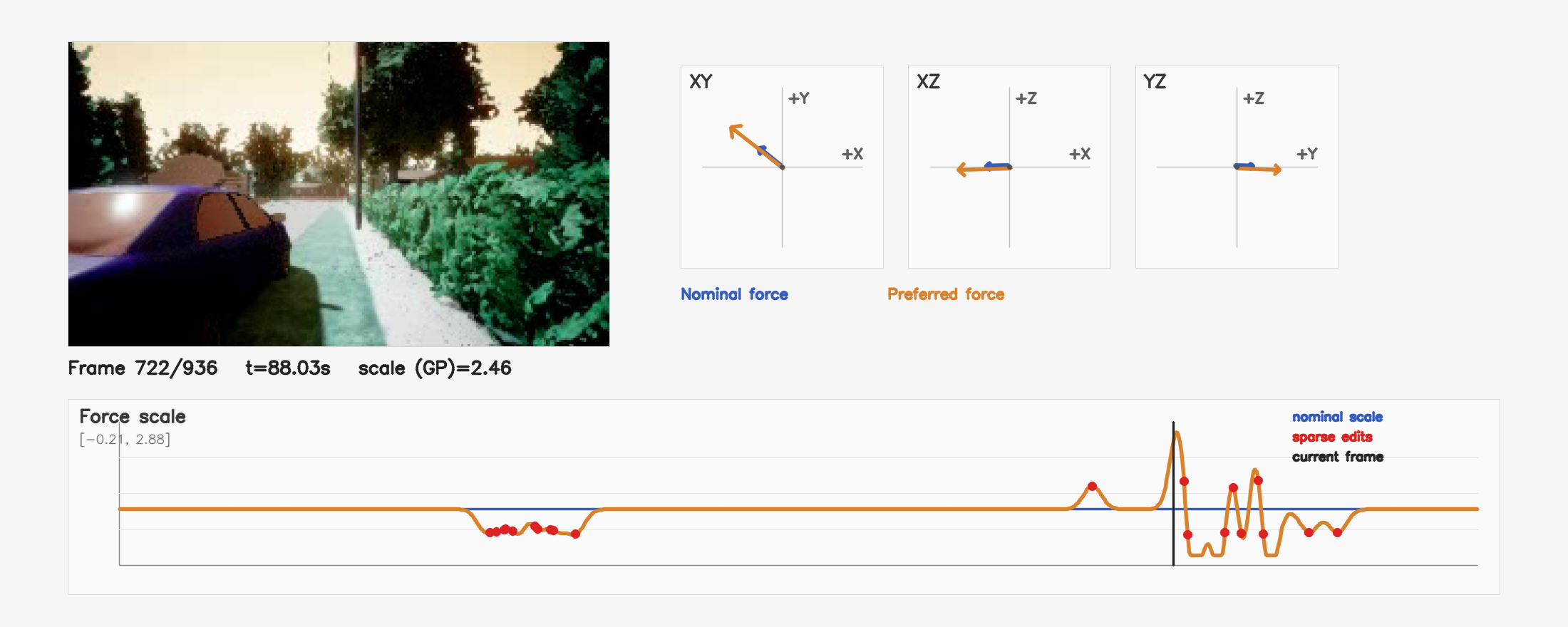}
    \caption{\textbf{Replay-based preference annotation interface.}
    The operator reviews the recorded FPV sequence, compares the original
    and modified haptic forces in three orthogonal projections, and sparsely
    adjusts the force magnitude while optionally replaying the modified force
    through the haptic device. The sparse demonstrations are interpolated using the GP procedure of \Cref{sec:GP} to construct the desired force trajectory used for learning.}
    \label{fig:gui}
\end{figure*}

An initial teleoperation session is performed using nominal CBF response gains $\Theta_0$. During the session, the operator commands the robot through the haptic interface, while the CBF-QP generates the haptic force $F_{\mathrm{cbf}}$ according to the current robot state and active safety constraints, as described in \Cref{sec:force feedback}. For each sample, we record $\mathcal{D}_{\mathrm{init}}$ and the nominal force ${\{F_{\mathrm{cbf,0}}^{(j)}\}_{j=1}^{N}}$. The synchronized FPV sequence is also recorded to provide visual context during preference specification.

Providing a desired force at every sample would impose a substantial burden on the operator. We therefore collect preferences through the replay-based interface illustrated in \Cref{fig:gui}. The operator reviews the recorded FPV sequence together with the generated haptic feedback, which is visualized by its projections onto the $xy$, $xz$, and $yz$ planes. In addition to the visual replay, the recorded haptic feedback is also rendered through the haptic device, allowing the operator to physically experience the generated intervention before refining it. The combination of visual and haptic replay enables the operator to evaluate and adjust the force feedback using both visual and kinesthetic information.

At selected time instants, the operator modifies only the magnitude of the recorded force while preserving its direction. This retains the corrective direction produced by the CBF-QP and allows the operator to specify how strongly the intervention should be communicated. Let $\rho_i$ denote the force-scaling factor specified by the operator at time instant $t_i$. The resulting sparse preference demonstrations are
\begin{equation}
    \mathcal{D}_{\mathrm{spar}}
    =
    \left\{
    (t_i,\rho_i)
    \right\}_{i=1}^{M},
    \qquad M \ll N.
    \label{eq:sparse_demonstrations}
\end{equation}
The corresponding preferred force at demonstration time $t_i$ is $\rho_iF_{\mathrm{cbf},0}(t_i)$.

To construct a desired force target over the complete trajectory, we apply
the GP regression introduced in \Cref{sec:GP} to the sparse scale
demonstrations. Specifically, we model the time-varying force scale
$\rho(t)$ using a GP with prior mean
\begin{equation}
    \mu_\rho(t)=1,
\end{equation}
so that the inferred preference returns to the nominal CBF response in regions not supported by user demonstrations. We use the radial basis function kernel
\begin{equation}
    k(t_a,t_b)
    =
    \exp\left(
        -\frac{(t_a-t_b)^2}{2r^2}
    \right),
    \label{eq:gp_scale_kernel}
\end{equation}
where $r>0$ is the kernel length scale that determines the temporal influence of each demonstration. The operator can adjust $r$ through the replay interface to control how broadly a modification propagates along the recorded trajectory. A smaller value of $r$ produces more localized changes and is useful when the preferred force varies rapidly, such as when transitioning between the car and nearby trees (as shown in the image frame in \Cref{fig:gui}). A larger value produces smoother modifications over a wider temporal interval.

Applying the GP regression to the sparse preference demonstrations $\mathcal{D}_{\mathrm{spar}}$ yields the posterior mean force scale $\hat{\rho}(t)$ and the posterior variance $\sigma_\rho^2(t)$. The posterior mean is evaluated at the recorded sample times to construct the desired force trajectory
\begin{equation}
    F_{\mathrm{wish}}^{(j)}
    =
    \hat{\rho}(t_j)
    F_{\mathrm{cbf},0}^{(j)},
    \qquad j=1,\ldots,N.
    \label{eq:desired_force_trajectory}
\end{equation}
The posterior variance quantifies the confidence of the interpolated preference and is used in the GP-weighted training variant.

Consequently, the operator specifies only sparse changes in intervention strength, while the GP converts these demonstrations into the smooth desired force trajectory $\mathcal{F}_{\mathrm{wish}}$ used in the subsequent parameter-learning stage. The posterior variance is low near the modified samples and increases away from them, indicating the confidence of the interpolated preference.

\subsection{Learning from Haptics}
\label{subsec:preference_learning}

\subsubsection{Learning Objective}
\label{subsubsec:learning_objective}

The learning objective is instantiated here for the grouped CBF response gains introduced in \Cref{sec:problem} and augmented with a regularization term. For each recorded sample $j$, the CBF-QP is reconstructed from $\mathcal{D}_{\mathrm{init}}$ and solved using the current parameter set $\Theta$. The resulting force $F_{\mathrm{cbf}}^{(j)}(\Theta)$ is compared with the desired force $F_{\mathrm{wish}}^{(j)}$ constructed in
\Cref{subsec:data_collection}. We define the sample-wise preference loss as
\begin{equation}
    \ell_j(\Theta)
    =
    \left\|
    F_{\mathrm{cbf}}^{(j)}(\Theta)
    -
    F_{\mathrm{wish}}^{(j)}
    \right\|_2^2.
    \label{eq:sample_preference_loss}
\end{equation}

The loss over the recorded trajectory is
\begin{equation}
    \mathcal{L}(\Theta)
    =
    \frac{1}{N}
    \sum_{j=1}^{N}
    w_j \ell_j(\Theta),
    \label{eq:weighted_learning_loss}
\end{equation}
where $w_j\geq 0$ denotes the sample weight. The choice of $w_j$
depends on the weighting strategy used during training. Uniform
weighting sets $w_j=1$ for all samples. Semantic-informativeness
weighting assigns larger weights to samples whose active constraints
are associated with the semantic groups most relevant to the
demonstrated corrections. GP-based weighting instead uses
$w_j=1/\bigl(\sigma_{\rho}^{2}(t_j)+\epsilon\bigr)$, where $\sigma_\rho^2(t_j)$ is the GP posterior variance and
$\epsilon>0$ prevents the weight from becoming unbounded.

To discourage excessive deviations from the nominal CBF response gains,
we include the regularization term
\begin{equation}
    \mathcal{R}(\Theta)
    =
    \lambda
    \left\|
    \Theta-\Theta_0
    \right\|_2^2,
    \label{eq:gain_regularization}
\end{equation}
where $\lambda\geq0$ controls the regularization strength. In practice, $\lambda$ is chosen to be small, so that the regularization primarily improves numerical stability without materially affecting the learned response gains. 

The resulting learning problem is
\begin{equation}
    \Theta^\star
    =
    \arg\min_{\Theta\in\Omega}
    \mathcal{L}(\Theta)
    +
    \mathcal{R}(\Theta),
    \label{eq:learning_objective}
\end{equation}
where $\Omega$ denotes the admissible set of CBF response gains. In particular, the gains are constrained to remain positive, ensuring that the characteristic polynomial $p_s(z)=z^2+k_{2,s}z+k_{1,s}$ is Hurwitz for each constraint group, and are bounded above to avoid excessively aggressive responses. The dependence of $F_{\mathrm{cbf}}^{(j)}(\Theta)$ on $\Theta$ is implicit through the solution of the CBF-QP. The following subsection describes how the gradients of \eqref{eq:learning_objective} are computed through the differentiable optimization layer.

\subsubsection{Backpropagation through the Differentiable CBF-QP}
\label{subsubsec:differentiable_optimization}
To optimize \eqref{eq:learning_objective}, we differentiate the CBF-QP solution with respect to the response gains. For clarity, we suppress the sample index $j$ in the following derivation. Given $m$ active CBF constraints, the QP in \eqref{eq:CBF_QP} can be written in matrix form as
\begin{subequations}
\label{eq:cbf_qp_matrix}
\begin{align}
    u_{\mathrm{cbf}}
    =
    \argmin_{u\in\mathbb{R}^{d}}
    \quad &
    \frac{1}{2}
    \left\|
    u-u_{\mathrm{ref}}
    \right\|_2^2
    \label{eq:cbf_qp_matrix_objective}
    \\
    \text{s.t.}\quad &
    A u \leq B(\Theta),
    \label{eq:cbf_qp_matrix_constraint}
\end{align}
\end{subequations}
where $A\in\mathbb{R}^{m\times d}$ and $B(\Theta)\in\mathbb{R}^{m}$. For constraint $i$, associated with group $s_i$, the corresponding row of $A$ and entry of $B(\Theta)$ are
\begin{align}
    A_i
    &=
    -\left(\partial_{x_p}h_i\right)^\top,
    \label{eq:cbf_matrix_A}
    \\
    B_i(\Theta)
    &=
    x_v^\top
    \partial_{x_p}^{2}h_i
    x_v
    +
    k_{2,s_i}
    \left(\partial_{x_p}h_i\right)^\top x_v
    +
    k_{1,s_i}h_i.
    \label{eq:cbf_matrix_b}
\end{align}
Thus, the recorded state and environment geometry determine $A$, while the grouped response gains enter the QP through $B(\Theta)$.

Let $u^\star$ denote the optimal QP solution and let $\mu^\star\in\mathbb{R}^{m}_{\geq 0}$ denote the associated dual variables. The relevant Karush--Kuhn--Tucker conditions are
\begin{equation}
\label{eq:cbf_qp_kkt}
\begin{aligned}
    u^\star-u_{\mathrm{ref}}
    +A^\top\mu^\star
    &=0,
    \\
    D(\mu^\star)
    \left(
    Au^\star-B(\Theta)
    \right)
    &=0,
\end{aligned}
\end{equation}
together with primal and dual feasibility, where $D(\cdot)$ constructs a diagonal matrix from its vector argument. Because $A$ is fixed for each recorded sample, differentiating \eqref{eq:cbf_qp_kkt} with respect to the response gains gives
\begin{equation}
\label{eq:cbf_qp_differential}
\begin{aligned}
    \mathrm{d}u^\star
    +A^\top\mathrm{d}\mu^\star
    &=0,
    \\
    D\!\left(
    Au^\star-B(\Theta)
    \right)
    \mathrm{d}\mu^\star
    +
    D(\mu^\star)
    \left(
    A\,\mathrm{d}u^\star
    -
    \mathrm{d}B
    \right)
    &=0.
\end{aligned}
\end{equation}
Solving this linear system provides the sensitivity of the optimal solution to perturbations in $B(\Theta)$ and, consequently, to the response gains in $\Theta$.

Using the force mapping defined in \eqref{eq:parameterized_force}, its differential is $\mathrm{d}F_{\mathrm{cbf}}=k_f\,\mathrm{d}u^\star$. These sensitivities are propagated through the sample-wise losses $\ell_j(\Theta)$ and aggregated to compute the gradient of
$\mathcal{L}(\Theta)$. In our implementation, the QP is represented as a differentiable optimization layer using CVXPYLayers \citep{agrawal2019differentiable}, while PyTorch performs backpropagation and updates the response gains. The admissible set $\Omega$ is enforced during optimization to maintain valid gain values. This procedure yields the learned parameter set $\Theta^\star$ in \eqref{eq:learning_objective}.

\subsection{Safety Constraint Construction}
\label{subsec:safety_constraint_construction}

The proposed learning framework is independent of a particular safety
representation. At each recorded sample, the constraint data
$\mathcal{C}^{(j)}$ contain the geometric quantities required to
reconstruct the active CBF constraints during learning.

For the double-integrator dynamics considered in this work, each active
constraint $i$ is represented by a barrier function $h_i(x_p)$ and its
spatial derivatives. The corresponding second-order CBF condition is
\begin{equation}
\begin{aligned}
    &
    x_v^\top
    \partial_{x_p}^{2}h_i
    x_v
    +
    \left(\partial_{x_p}h_i\right)^\top u
    \\
    &\qquad
    +
    k_{2,s_i}
    \left(\partial_{x_p}h_i\right)^\top x_v
    +
    k_{1,s_i}h_i
    \geq 0,
\end{aligned}
\label{eq:representation_general_cbf}
\end{equation}
where $s_i\in\mathcal{S}$ denotes the group associated with constraint
$i$. Therefore, a safety representation can be incorporated into the
proposed framework as long as it provides $h_i$,
$\partial_{x_p}h_i$, and $\partial_{x_p}^{2}h_i$.

The resulting quantities determine the corresponding row of the matrix-form CBF-QP in \eqref{eq:cbf_qp_matrix_constraint}. The environment representation therefore specifies the geometry of the intervention, while the learned response gains determine how strongly and how early the intervention is communicated to the operator.  The specific FPV perception-based and signed distance field (SDF)-based constraint constructions
are described in the corresponding simulation and hardware validation
sections.

\section{Validation}\label{sec:experiments}
We evaluate the proposed framework in two aerial HSC settings. The AirSim hardware-in-the-loop experiment provides a controlled and repeatable setting for quantitatively evaluating how well the learned CBF response gains reproduce the demonstrated haptic preferences. It uses FPV perception-based semantic constraints and allows the nominal, demonstrated, and learned force trajectories to be compared over the same recorded teleoperation session.

The hardware experiment evaluates whether the learned response gains can be deployed in a physical aerial teleoperation loop. A DJI Tello drone is controlled through a Geomagic Touch device, while SDF-based CBF constraints are constructed from motion-capture measurements. Because the nominal and personalized responses are obtained from separate manually controlled flights, this experiment is presented as a qualitative deployment demonstration rather than a frame-wise force-matching evaluation.

The two experiments evaluate the quantitative preference-reproduction performance of the proposed learning method, its deployment in a physical human-robot interaction loop, and its compatibility with different constructions of the underlying CBF constraints.

\subsection{AirSim Simulation Evaluation}
\label{subsec:airsim_validation}

\subsubsection{Experimental Setup}
\label{subsubsec:airsim_setup}
\begin{figure}[t]
    \centering
    \includegraphics[width=\linewidth]{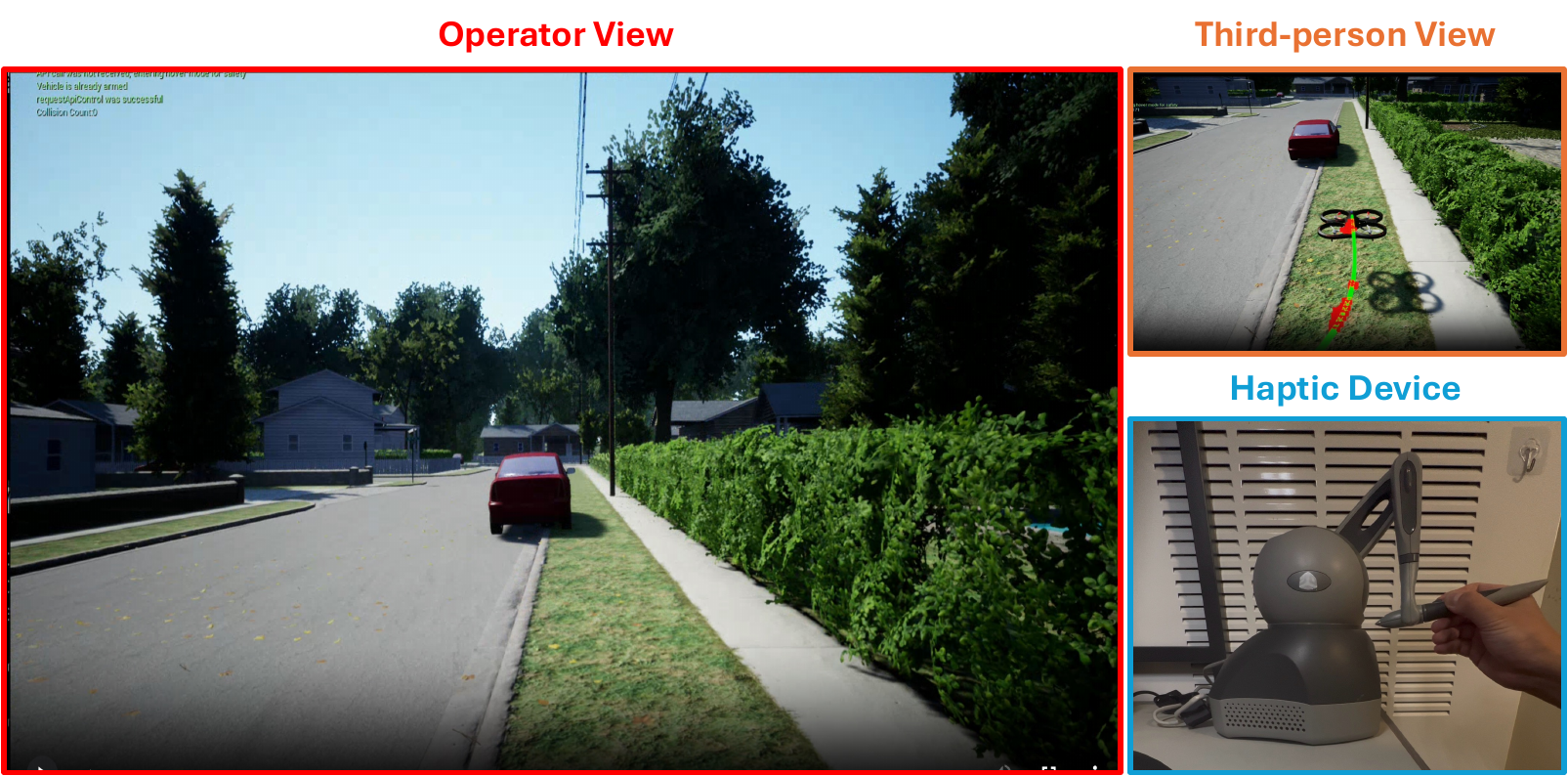}
    \caption{\textbf{AirSim hardware-in-the-loop simulation setup.}
    The operator commands the simulated drone and receives CBF-generated haptic feedback through the Geomagic Touch device. Only the first-person view is presented during teleoperation.}
    \label{fig:simulation_setup}
\end{figure}

The AirSim ~\citep{shah2018airsim} hardware-in-the-loop setup is shown in \Cref{fig:simulation_setup}. The simulated aerial robot is teleoperated using a Geomagic Touch haptic device. The displacement of the haptic stylus from its neutral position is mapped to the desired drone velocity $x_{vd}$, from which the reference acceleration $u_{\mathrm{ref}}$ is computed using~\eqref{eq:uref}.

During both the initial recording and personalized deployment, the operator receives only the first-person camera view shown in \Cref{fig:simulation_setup}. The third-person view is included in the figure solely to illustrate the simulated environment and is not available to the operator during teleoperation. AirSim provides synchronized RGB, depth, semantic labels, and robot-state information for constructing the perception-based CBF constraints described in \Cref{subsubsec:fpv_constraints}.

An initial teleoperation session is performed using the nominal response gains $\Theta_0$. During this session, the robot states, user reference inputs, active constraint data, nominal haptic forces, and synchronized FPV frames are recorded. The operator subsequently reviews the recorded session through the replay interface and provides sparse modifications of the force magnitude. These annotations are interpolated to construct the desired force trajectory used for learning.

The grouped differentiable CBF-QP is optimized for 120 epochs using 358 informative samples from the recorded session. A sample is considered informative when at least one perception-derived constraint is active and the nominal CBF-QP produces a nonzero haptic response. Samples that do not satisfy these conditions provide limited information
about the response gains and are therefore excluded from the reported force-matching metrics. Unless otherwise stated, all quantitative AirSim results are evaluated over the same set of informative samples.

\subsubsection{FPV Perception-Based Constraints}
\label{subsubsec:fpv_constraints}

At each simulation step, AirSim provides synchronized RGB, depth,
and semantic-label images from the onboard camera. A central region
of interest (ROI) is selected from the FPV, and the
valid pixels within the ROI are grouped according to their semantic
labels. Each visible semantic group may generate one active safety
constraint.

For constraint $i$, let $\mathcal{P}_i$ denote the valid pixels
assigned to the corresponding semantic group. A representative depth
$d_i$ is computed using a depth percentile rather than the minimum
depth to reduce sensitivity to isolated measurements. Let $p_i$
denote the closest valid pixel in $\mathcal{P}_i$, and let
$\bar{r}_i$ be its corresponding normalized camera ray. The
local clearance-increasing direction is approximated as
\begin{equation}
    n_i=-R_{WC}\bar{r}_i,
    \label{eq:fpv_normal}
\end{equation}
where $R_{WC}$ denotes the camera-to-world rotation.

The barrier value is defined as
\begin{equation}
    h_i=d_i-d_{\mathrm{safe}},
    \label{eq:fpv_barrier}
\end{equation}
where $d_{\mathrm{safe}}>0$ is the prescribed minimum clearance.
Treating $n_i$ as locally constant during one control update
gives
\begin{equation}
    \dot h_i\approx n_i^\top x_v,
    \qquad
    \ddot h_i\approx n_i^\top u.
    \label{eq:fpv_barrier_dynamics}
\end{equation}
The resulting CBF constraint is
\begin{equation}
    n_i^\top u
    +k_{2,s_i}n_i^\top x_v
    +k_{1,s_i}h_i
    \geq 0,
    \label{eq:fpv_cbf_constraint}
\end{equation}
where $s_i$ denotes the semantic group associated with constraint $i$. Constraints in the same group share the corresponding response gains. This construction generates local semantic safety constraints
directly from the current FPV observation without requiring an analytical obstacle model or a global map.

\subsubsection{Force Preference Matching}
\label{subsubsec:force_matching}

\begin{figure*}[t]
    \centering
    \includegraphics[width=\linewidth]
    {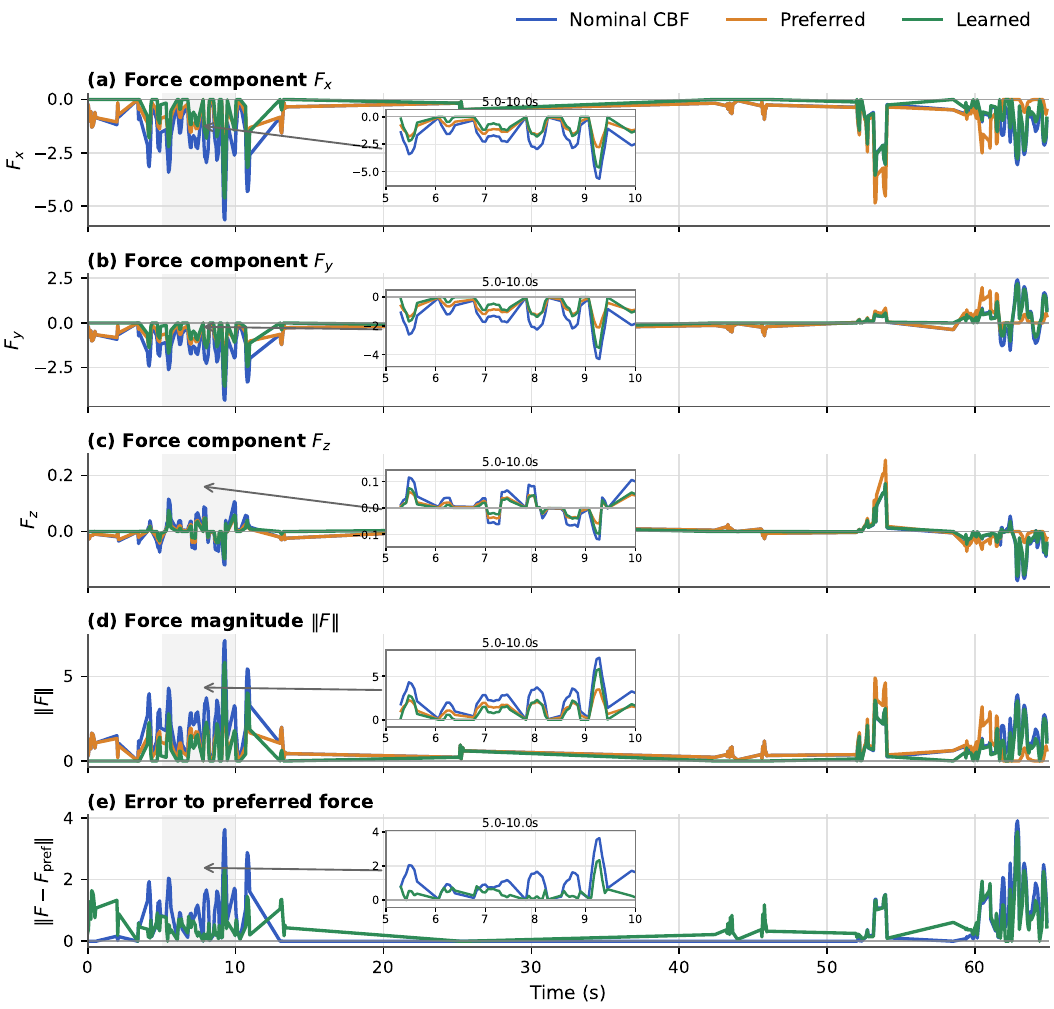}
    \caption{\textbf{Force preference matching over the recorded AirSim
    teleoperation session.}
    The nominal CBF force, demonstrated force preference, and force
    regenerated using the learned grouped gains are compared. Panels
    (a)--(c) show the force components, panel (d) shows the force
    magnitude, and panel (e) shows the instantaneous Euclidean error
    relative to the demonstrated preference.}
    \label{fig:force_comparison_overview}
\end{figure*}

We evaluate whether the proposed learning framework reduces the
mismatch between the generated haptic force and the demonstrated
preference. \Cref{fig:force_comparison_overview} compares the force
trajectories over the recorded teleoperation session, while
\Cref{tab:force_matching_results} reports the aggregate force-matching
errors for different parameterizations and sample-weighting strategies.

\begin{table}[t]
\centering
\caption{Force preference matching over the recorded AirSim FPV
teleoperation session. Lower values indicate closer agreement with
the demonstrated preference.}
\label{tab:force_matching_results}
\begin{tabular}{lcc}
\hline
Method & Mean Error & MSE \\
\hline
Nominal CBF & 0.8215 & 1.8786 \\
Global DiffQP & 0.7514 & 1.5147 \\
Top-$K$ grouped & 0.6859 & 1.4451 \\
Top-$K$ semantic-weighted & 0.6857 & 1.4457 \\
Top-$K$ GP-weighted & 0.6904 & 1.5747 \\
\hline
\end{tabular}
\end{table}

The nominal CBF baseline uses manually selected response gains without
learning. Global DiffQP learns one gain pair shared by all active
constraints through the differentiable CBF-QP layer. The Top-$K$
variants learn separate gain pairs for the $K$ semantic groups identified
as the most informative based on the recorded preference corrections,
while the remaining groups retain their nominal gains. In this experiment,
we use $K=2$. Top-$K$ semantic-weighted further weights the training
samples according to the semantic informativeness of the active
constraints, whereas Top-$K$ GP-weighted assigns larger weights to samples with lower GP
posterior variance.

Mean Error is the average Euclidean distance between the generated force
and the demonstrated preference over the informative samples, and MSE is
the corresponding mean squared error. As shown in
\Cref{fig:force_comparison_overview}, the force regenerated using the
learned gains follows the demonstrated preference more closely than the
nominal response, particularly during intervals in which the operator
modifies the intervention strength.

The results in Table~\ref{tab:force_matching_results} show that learning
a single global gain pair reduces the MSE from $1.8786$ to $1.5147$,
corresponding to a $19.4\%$ reduction relative to the nominal CBF.
Introducing semantic-specific gains further reduces the MSE to $1.4451$,
which represents a $23.1\%$ reduction relative to the nominal response
and a further $4.6\%$ reduction relative to Global DiffQP. This result
indicates that the grouped parameterization better captures variations
in the demonstrated intervention preference. The semantic-weighted
variant performs similarly to the uniformly weighted grouped model. The
GP-weighted variant remains better than the nominal baseline but produces
a higher MSE than the other Top-$K$ variants.

Overall, the results suggest that semantic grouping has a greater effect on force preference matching in this recording than the choice of sample-weighting strategy. The lower performance of GP-based weighting may result from placing excessive emphasis on samples near the sparse edits, thereby reducing the influence of the remaining portions of the force trajectory. A residual mismatch remains because the demonstrated preference contains frame-dependent modifications, whereas the learned model represents them using a compact set of gains shared within each semantic group.

\subsection{Hardware Demonstration}
\label{subsec:hardware_demo}

\subsubsection{Experimental Setup and Safety Representation}
\label{subsubsec:hardware_setup}
The hardware study evaluates the deployment of the proposed learning
framework in a physical aerial teleoperation loop and examines its compatibility with a safety representation different from the FPV perception-based method. A DJI Tello drone is teleoperated
using the Geomagic Touch device, while the drone and obstacle states are measured by an OptiTrack motion-capture system. These measurements are used to construct SDF-based CBF constraints.

The experimental environment contains one rectangular obstacle and two circular obstacles. Communication between the drone, host computer, and haptic device is implemented through ROS and Tello's local WiFi network. 

\begin{figure}[t]
    \centering
    \includegraphics[width=0.95\columnwidth]
    {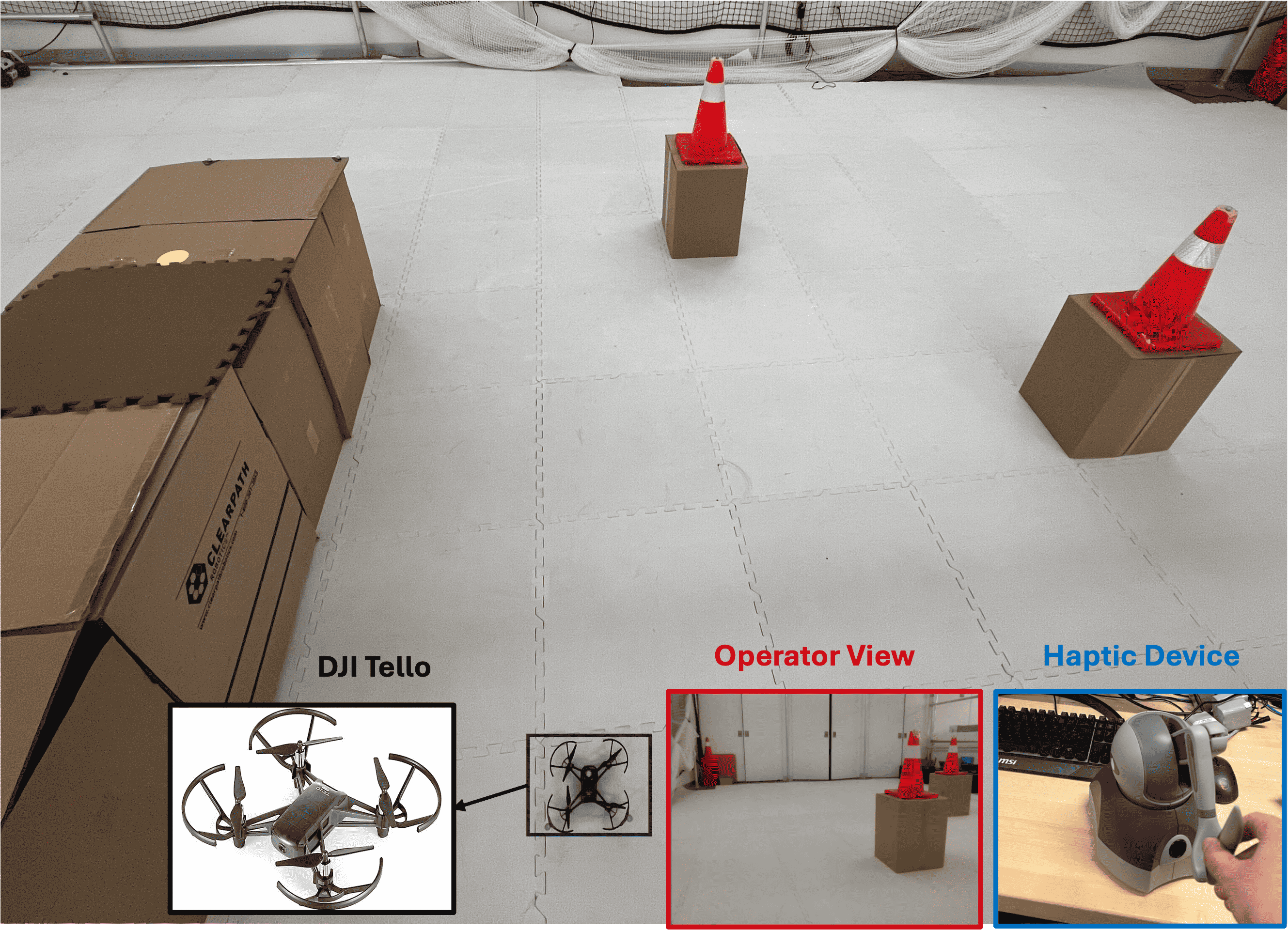}
    \caption{\textbf{Hardware teleoperation setup.}
    A Geomagic Touch device is used to command the DJI Tello EDU drone
    and render haptic safety interventions. The operator receives the
    synchronized first-person camera view, while OptiTrack measurements
    provide the drone and obstacle poses used to construct the SDF-based
    safety constraints.}
    \label{fig:exp_setup}
\end{figure}

We use $d_{\mathrm{SDF},i}(x_p)$ for the signed distance associated with
obstacle $i$. In practice, the SDF is represented on a discrete grid over the workspace. The signed distance values are computed offline from the obstacle geometry or from an occupancy representation using a distance transform procedure. During control, the SDF value at the robot position is obtained by interpolation from neighboring grid values. The gradient and Hessian are computed from the discretized SDF using finite differences and are interpolated at the current robot position. This provides the barrier value, local surface normal direction, and second-order curvature information required by the CBF constraint without solving an online geometric optimization problem.
The corresponding barrier function is
\begin{equation}
    h_i(x_p)
    =
    d_{\mathrm{SDF},i}(x_p)
    -
    d_{\mathrm{safe}},
    \label{eq:sdf_barrier}
\end{equation}
and its spatial derivatives are
\begin{equation}
    \partial_{x_p}h_i
    =
    \nabla d_{\mathrm{SDF},i}(x_p),
    \qquad
    \partial_{x_p}^{2}h_i
    =
    \nabla^2 d_{\mathrm{SDF},i}(x_p).
    \label{eq:sdf_derivatives}
\end{equation}
These quantities are inserted into the same second-order CBF-QP used in
the AirSim study. Each obstacle is assigned a separate constraint group,
allowing its response gains to be adapted independently through the same
replay-based preference learning procedure.

\subsubsection{Personalized Hardware Intervention}
\label{subsubsec:hardware_results}

The initial teleoperation session is performed using nominal CBF response gains. The operator subsequently reviews the recorded FPV and specifies a stronger intervention near obstacle A, no intervention associated with obstacle B, and a weaker intervention near obstacle C.
The resulting demonstrations are used to learn obstacle-specific CBF response gains.

\begin{figure}[t]
    \centering
    \subfloat[Nominal response before learning.
    \label{subfig:before_learning}]
    {\includegraphics[width=\columnwidth]
    {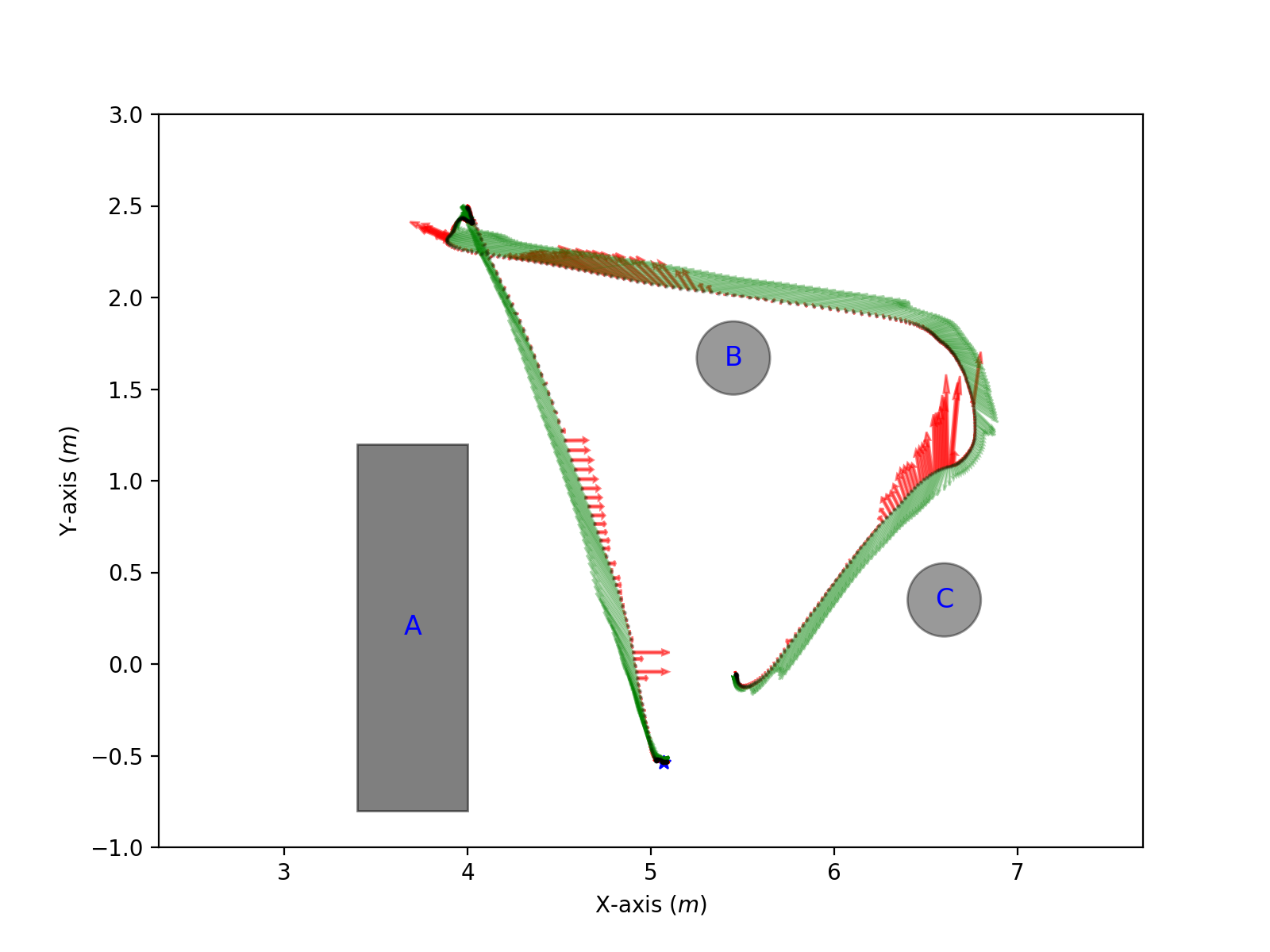}}
    \\
    \subfloat[Personalized response after learning.
    \label{subfig:after_learning}]
    {\includegraphics[width=\columnwidth]
    {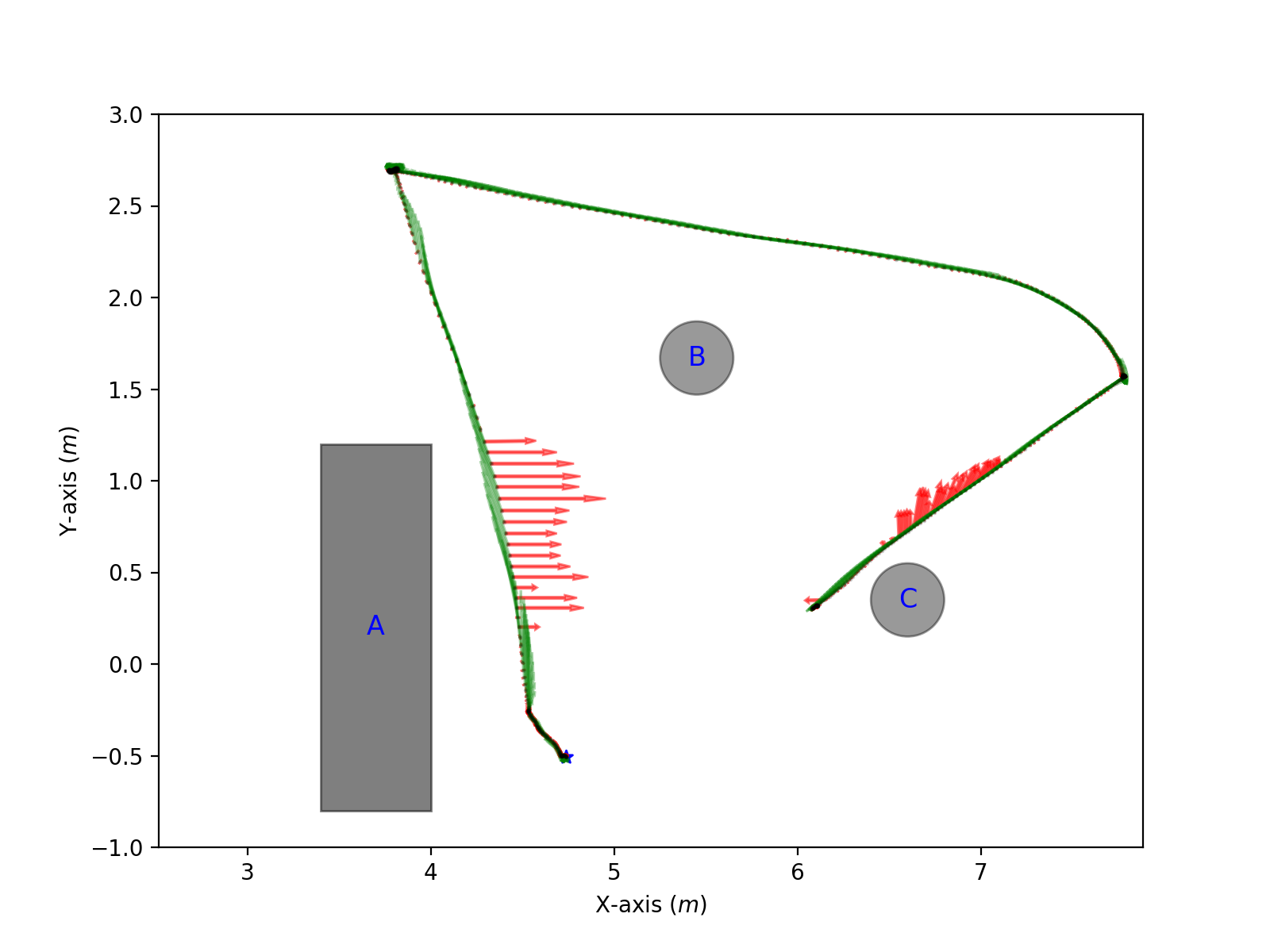}}
    \caption{\textbf{Hardware haptic intervention before and after
    learning.}
    Black curves show the drone trajectories, blue stars indicate the
    initial positions, green arrows show the drone velocity, and red
    arrows visualize the rendered haptic force. The learned response
    produces obstacle-dependent intervention strengths consistent with
    the operator's demonstrated preferences.}
    \label{fig:exp_result}
\end{figure}

The nominal response is shown in \Cref{subfig:before_learning}, while \Cref{subfig:after_learning} shows the intervention generated with the learned gains. After learning, the rendered force becomes stronger near obstacle A, is suppressed near obstacle B, and remains weaker near obstacle C, consistent with the demonstrated preference.

Because the drone is manually teleoperated from its FPV, the operator cannot reproduce identical state and command trajectories across separate flights. The two plots should therefore be interpreted as a deployment demonstration rather than a frame-wise force comparison. The experiment shows that the learned obstacle-specific response gains can be deployed with a physical drone and a haptic device, and that the framework is compatible with SDF-based constraints.

\section{Conclusions and Future Work}

This paper presented a Learning from Haptics framework for personalizing CBF-based interventions in haptic human-robot shared control. The framework allows users to specify their preferred intervention behavior through sparse modifications of replayed haptic feedback, avoiding direct manual tuning of the underlying CBF response gains. Gaussian process regression constructs a continuous desired force trajectory from the sparse demonstrations, and a differentiable CBF-QP learns grouped response gains that approximate the demonstrated preference.

The AirSim evaluation showed that the learned gains reduced the force mismatch relative to the nominal CBF response. Learning semantic-specific gain pairs further improved preference matching compared with learning a single global gain pair. The hardware experiment demonstrated deployment with a physical drone and produced obstacle-dependent intervention behaviors consistent with the specified preferences. Together, the FPV perception-based simulation and SDF-based hardware experiments show that the proposed learning formulation is compatible with different constructions of the underlying CBF constraints.

Future work will investigate nonlinear and learning-based parameterizations of the CBF response, as well as generalization across trajectories, environments, and operating conditions. A comprehensive user study will evaluate annotation effort, perceived workload, and the usability of the preference-learning procedure.



\section*{Declarations}
\paragraph{Conflict of interest} The authors have no relevant financial or non-financial
interest to disclose.

\paragraph{Availability of data}
The datasets generated and/or analyzed during the current study are available from the corresponding author upon reasonable request. 

\paragraph{Ethics approval}
The authors have no relevant ethics approval to disclose.

\paragraph{Funding information}
This work was funded by the Division of Systems Engineering at Boston University.

\paragraph{Author contributions}
D.Z. and R.T. conceptualized the study, designed the methodology, performed the experimental validation, and wrote the manuscript. R.T. supervised the study and contributed to the scientific discussions, data interpretation, and the manuscript.

\paragraph{}

\bibliography{biblio/IEEEfull,biblio/IEEEConfFull,biblio/OtherFull,biblio/dawei,biblio/controlLearningCBF,biblio/learning_cbf}

\end{document}